\documentclass[10pt,twocolumn,letterpaper]{article}

\usepackage{cvpr}              
\definecolor{cvprblue}{rgb}{0.21,0.49,0.74}
\definecolor{HeaderGray}{gray}{0.90}   %
\definecolor{StripeGray}{gray}{0.96}   %
\definecolor{SpiderBlue}{RGB}{220,236,250}
\usepackage[pagebackref,breaklinks,colorlinks,allcolors=cvprblue]{hyperref}

\usepackage{colortbl}
\usepackage{multirow}
\usepackage{bm}
\usepackage{makecell}
\usepackage{graphicx}
\usepackage{pifont}
\usepackage{lineno}
\usepackage[dvipsnames]{xcolor}
\usepackage{wrapfig}
\usepackage{caption}
\usepackage{subcaption}
\usepackage{marvosym}
\usepackage[table]{xcolor}
\usepackage{arydshln} 

\newcommand\blfootnote[1]{%
  \begingroup
  \renewcommand\thefootnote{}\footnote{#1}%
  \addtocounter{footnote}{-1}%
  \endgroup
}

\def\paperID{4709} 
\def\confName{CVPR}
\def\confYear{2026}

\title{DEGround: An Effective Baseline for Ego-centric 3D Visual \\ Grounding with a Homogeneous Framework}

\begin{document}
\author{
Yani Zhang$^{1*\dagger}$,
Dongming Wu$^{2*}$,
Hao Shi$^{3}$,
Yingfei Liu$^{4\ddagger}$,
Tiancai Wang$^{4}$,
Xingping Dong$^{1}$\textsuperscript{\Letter}\\
$^1$ School of Computer Science, Wuhan University \\
$^2$ MMLab, The Chinese University of Hong Kong, \\
$^3$ Tsinghua University,
$^4$ Dexmal \\
{\tt\small \{zebrazyn, xinpngdong\}@whu.edu.cn,}
{\tt\small wudongming97@gmail.com}
}
\maketitle
\blfootnote{$^*$Equal contribution. \textsuperscript{\Letter}Corresponding author: \textit{Xingping Dong}.
$^\dagger$This work was done during the internship at Dexmal.
$^\ddagger$Project lead.
This work was supported in part by the New Generation Artificial Intelligence-National Science and Technology Major Project (No. 2025ZD0123501), the National Natural Science Foundation of China under Grant 62471342, and WHU-Kingsoft Joint Lab.
}

\begin{abstract}
A core task in embodied intelligence is ego-centric 3D visual grounding. Existing methods typically adopt two-stage, heterogeneous pipelines that pair a detector with a separate grounding model.
Incompatible decoders and box heads hinder the transfer of object-level priors, and the split training causes redundant re-optimization. 
To overcome these limitations, we present \textbf{DEGround}, a straight, elegant, and effective framework that centers on object-level sharing over detection and grounding. 
It employs a set of queries that serves as the common object representation for both detection and grounding, which is decoded by a shared transformer and bounding box head.
Building on this homogeneous framework, we further introduce two task-specific plug-in modules to enhance fine-grained instruction grounding.
The Regional Activation Grounding module improves spatial-textual alignment by highlighting instruction-relevant regions, while the Query-wise Modulation module applies sentence-conditioned affine modulation to generate instruction-aware queries at initialization.
Extensive experiments demonstrate that DEGround achieves the best performance on multiple benchmarks. 
Remarkably, it significantly outperforms previous methods
by \textbf{7.52\%} at overall precision on the EmbodiedScan dataset.
\end{abstract}    
\section{Introduction}
\label{sec:intro}
3D visual grounding (3D VG) plays an important role in Embodied AI, which aims to localize the object referred to by natural language within 3D scenes~\cite{chen2020scanrefer,zhang2023multi3drefer,achlioptas2020referit3d,zhang2024bootstrapping}.
Early approaches~\cite{huang2021text,yuan2021instancerefer,wu2024rg} often assume access to pre-reconstructed scene-level priors, but this contradicts the perceptual limitations of embodied agents in reality. 
Consequently, increasing attention has begun to focus on ego-centric 3D visual grounding~\cite{zhengdensegrounding,peng2025proxytransformation}, where agents need to reason based on sparse and partially visible first-person multi-view RGB-D observations. 
To standardize evaluation in this setting, EmbodiedScan~\cite{wang2024embodiedscan} provides a series of benchmarks for ego-centric 3D perception, which has catalyzed a series of subsequent methods~\cite{zheng2024denseg,zhengdensegrounding,peng2025proxytransformation,liang2024spatioawaregrouding3d}.
\begin{figure}
	\centering
	\includegraphics[width=\linewidth]{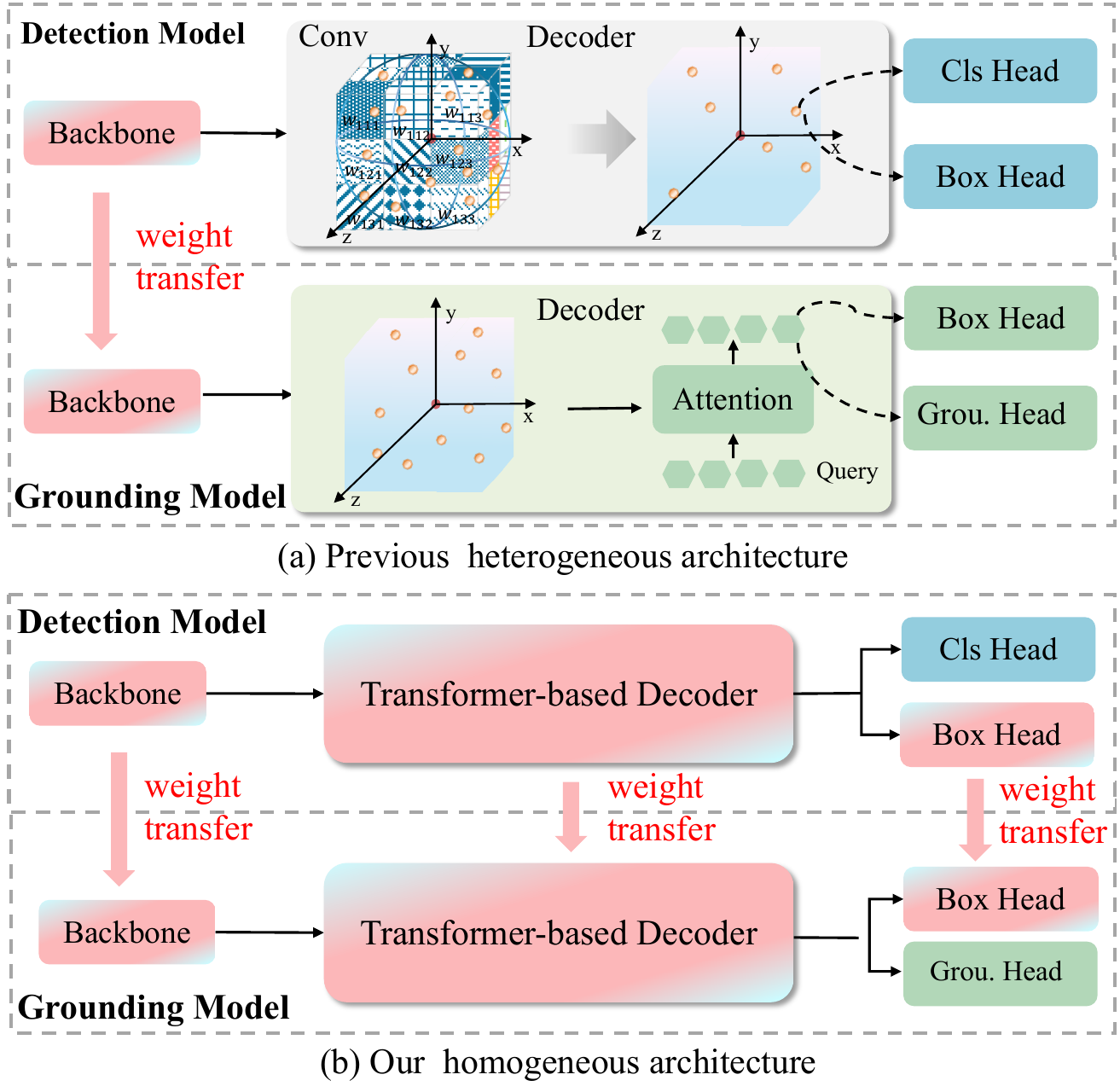}
	\vspace{-20pt}
	\caption{Comparison of past heterogeneous architecture and our homogeneous architecture.
    \textbf{Top}: Prior heterogeneous pipelines shares only backbone weight, leaving decoders/heads task-specific and supervision paths disjoint. \textbf{Bottom}: We treat object queries as a common representation and share the transformer decoder and box head between detection and grounding. This aligns supervision at the object level, enabling object-level prior transfer, faster convergence and superior performance.
    }
	\label{fig:motivation}
	\vspace{-18pt}
\end{figure} 

In ego-centric 3D VG, most previous works~\cite{wang2024embodiedscan,zheng2024denseg,zhengdensegrounding,peng2025proxytransformation,liang2024spatioawaregrouding3d} adopt a two-stage training scheme: a detector is first trained, and the resulting weights are subsequently transferred to initialize the grounding model.
This design aims to alleviate the difficulty of cross-modal alignment in grounding tasks by introducing stronger visual priors before joint vision-language learning.
As illustrated in~\cref{fig:motivation} (a), such transfer typically occurs at the backbone level, since the decoder and bounding-box head are architecturally heterogeneous across detection and grounding.
However, this heterogeneity leads to two main limitations:
1) \textit{Incomplete transfer of object-level priors.}
The backbone captures scene-level context. However, the object-level knowledge, which comprises the object representations from the decoder and the geometric reasoning refined by the bounding-box head, is not transferred. This prevents the grounding model from reusing the detector's learned object priors.
2) \textit{Redundant learning}. Even with detector initialization, the grounding model must re-learn the object decoding of instances and the bounding-box regression on the same scenes, incurring additional iterations and data passes and resulting in slow convergence and suboptimal accuracy.

To address this gap, we propose a novel homogeneous framework for \textbf{DE}tection and \textbf{Ground}ing, termed as DEGround.
Its central principle is object-level knowledge transfer, which is realized through a sharing query strategy. We maintain a single set of queries that serves as the common object representation for the two tasks. 
These queries are processed by a transformer-based decoder and a bounding-box head whose architecture and parameters are shared across tasks, as shown in~\cref{fig:motivation} (b). 
In contrast to the heterogeneous architecture in~\cref{fig:motivation} (a), where sharing is limited to the backbone, DEGround extends knowledge sharing to the object level, enabling the transfer of object representation and geometric reasoning. This alignment at the object level not only preserves and enhances detector priors but also reduces re-optimization overhead, leading to faster convergence and higher data efficiency. 
As discussed in~\cref{sec:experiments}, on the mini set, adding sharing query to the baseline yields 59.90 mAP within only 3 epochs, whereas the original baseline requires 12 epochs to attain 35.84 mAP.

Building on the homogeneous framework, we further enhance grounding effectiveness by incorporating two plug-in modules that can comprehend fine-grained instructions. 
The core idea behind them is to seamlessly refine the visual feature map and query space to align with grounding instructions as well as inherit detection sharing architecture.
First, we design a Regional Activation Grounding (RAG) module to handle same-class distractor cases, where objects share category labels but do not align with the language description. 
It highlights language-relevant regions via text-guided attention over the 3D feature map, while a spatial relevance loss provides point-level supervision from bounding box annotations.
As a result, RAG delivers a language-conditioned feature map to the subsequent decoder.
Second, we propose a Query-wIse Modulation (QIM) module, which injects linguistic information into the initial object queries to make them instruction-aware. 
It applies a feature-wise affine transformation to each query, with modulation parameters generated from the sentence embedding. This process adapts the query representation space to better integrate visual evidence with linguistic intent, thereby improving query discriminability from the initial stage of decoding.

Following~\cite{zhengdensegrounding}, we conduct extensive experiments on the EmbodiedScan~\cite{wang2024embodiedscan} benchmark, which aggregates data from ScanNet~\cite{scannet}, 3RScan~\cite{wald2019rio}, and Matterport3D~\cite{chang2017matterport3d} with more comprehensive annotations. This benchmark comprises multiple sub-benchmarks, covering ego-centric 3D understanding. 
DEGround consistently outperforms prior methods across these benchmarks, validating the effectiveness of our homogeneous framework and cross-modal modules.
Notably, for ego-centric 3D visual grounding, it achieves a significant improvement of \textbf{7.52\%} on the challenging full validation set. 
It also sets a new state-of-the-art performance on the official test set of the grounding task.

In summary, our main contributions are four-fold: 

\begin{itemize}
    \item We present DEGround, which uses sharing query as object representation for both tasks, achieving object-level sharing and removing redundant optimization.
    \item We introduce a novel regional activation grounding module that effectively highlights instruction-consistent regions with point-wise supervision. 
    \item We propose the query-wise modulation module that integrates global sentence-level semantics into the query representations, producing instruction-aware queries.
    \item We conduct extensive experiments on the embodied perception benchmark EmbodiedScan. DEGround outperforms existing methods by a large margin, resulting in an improvement of \textbf{7.52\%} over the state-of-the-art model.
\end{itemize}
\section{Related Work}
\subsection{3D Object Detection}
3D object detection serves as a foundational task for embodied perception. 
Early 3D detectors~\cite{shi2019pointrcnn,shi2020pv,lang2019pointpillars,rukhovich2022fcaf3d,ding2019votenet} largely rely on anchor-based or proposal-based paradigms that operate on point clouds or voxelized representations. For example, PointRCNN~\cite{shi2019pointrcnn} encodes strong geometric priors but depends on dense proposals and hand-crafted heuristics. 
More recently, transformer-based detectors~\cite{kolodiazhnyi2025unidet3d,shen2023v,wang2023uni3detr,misra2021end} build on query–feature interaction to dispense with dense anchors. 
Representative efforts include 3DETR~\cite{misra2021end} and Group-Free 3D~\cite{liu2021group} that adapt DETR-style set prediction to raw point clouds.

In ego-centric 3D VG, detectors are often adopted as the first-stage model to acquire spatial and categorical priors, which are later transferred to guide the grounding model. 
However, classical 3D detection commonly relies on point-based dense prediction, whereas grounding models adopt an object-centric, DETR-style set prediction with a small number of queries.
This architectural heterogeneity changes supervision pathways and results in weak transfer of priors across stages.
To bridge this divide, we introduce DEGround, a homogeneous DETR-style framework that enables seamless prior transfer.

\subsection{3D Visual Grounding}
3D visual grounding plays a vital role in robotic vision perception~\cite{deitke2022️,tian2023occ3d,luo20223d,caesar2020nuscenes,chen2020soundspaces,wu2023referring,wu2025language}.
Its core concept involves identifying and localizing target objects within 3D visual scenes based on natural language descriptions. 
A variety of datasets~\cite{achlioptas2020referit3d,chen2020scanrefer,azuma2022scanqa,zhang2023multi3drefer} have been introduced to advance progress in this field. In particular, ReferIt3DNet~\cite{achlioptas2020referit3d} and ScanRefer~\cite{chen2020scanrefer} established point-centric baselines and evaluation protocols.
Prior approaches primarily employ two-stage methods~\cite{chen2020scanrefer,huang2021text,he2021transrefer3d,yuan2021instancerefer,zhao20213dvg,yang2021sat,feng2021free,jain2022bottom,chen2022language,wu2024rg}, 
where pre-trained object detectors are first applied to generate object proposals, and the target is subsequently selected from these proposals by designing various matching strategies.
Building on this paradigm, LanguageRefer~\cite{chen2022language} leverages a language model to jointly encode proposals and queries, using it to estimate proposal confidence. Despite their effectiveness, two-stage methods suffer from detector bias and computational overhead from processing object proposals.

In addition to these two-stage models, single-stage
grounding approaches have emerged in the 3D domain~\cite{mvt,guo2023viewrefer,hsu2023ns3d,chang2024mikasa,shi2024aware,unal2024four,xu2024multi,zhang2024towards,wang2024g}.
For instance, MVT~\cite{mvt} employs a technique of rotating 3D scenes from various angles and encoding them to generate multi-view representations for bounding box prediction. 
While a previous study~\cite{deng2024can} explores the influence of natural language instructions, its focus is primarily on variations in language style, such as different accents or tones.
In contrast, our work delves into studying the effectiveness of language instructions themselves.
Moreover, most existing methods primarily focus on enhancing scene understanding using \textit{reconstructed scene-level} 3D point clouds or meshes.
However, scene-level information is not readily available in real-world applications.
In this work, we target the more challenging setting of ego-centric 3D visual grounding. The model consumes sensor-generated 3D inputs from first-person multi-view RGB-D, and it does not rely on ground-truth object proposals. This yields sparser observations and a more faithful match to the perceptual constraints of agents.

\subsection{Multi-task learning with 3D Grounding}
Leveraging 3D grounding and multi-task learning within a single framework has emerged as a highly active and trending research topic in the embodied agent community.
Most existing methods employ LLM-based frameworks~\cite{huang2023embodied,huang2023chat,hong20233d,chen2024grounded,zheng2024towards,wu2025ragnet,sun2025spacevista,sun2026autofly}, which convert inputs and outputs into visual or linguistic tokens to enable multi-task learning.
For example, LEO~\cite{huang2023embodied} collects large-scale 3D vision-language datasets and builds an embodied multi-modal generalist agent that excels in perceiving, grounding, reasoning, planning, and acting in the 3D world.
However, these methods still rely on an off-the-shelf 3D mask or box proposal module, requiring feeding these proposals into an LLM for grounding prediction. 
Different from them, PQ3D~\cite{zhu2024unifying} introduces promptable queries to unify multiple 3D vision–language tasks. 
However, PQ3D does not unify detection with language understanding and relies on pre-stored boxes.
BIP3D~\cite{lin2024bip3d} extends Grounding-DINO~\cite{liu2024grounding} by incorporating image features with 3D position encoding. 
However, this image-centric method lacks explicit 3D space modeling and relies heavily on extensive category label inputs for its detection task.
This intensive use of image data leads to high computational cost and slow inference.
In this work, we propose a simple yet powerful framework designed for ego-centric 3D visual grounding, which does not rely on pre-stored boxes and unifies detection and grounding object representation through sharing query.

\section{Method}
\label{sec:method}
\begin{figure*}
	\centering
	\includegraphics[width=\linewidth]{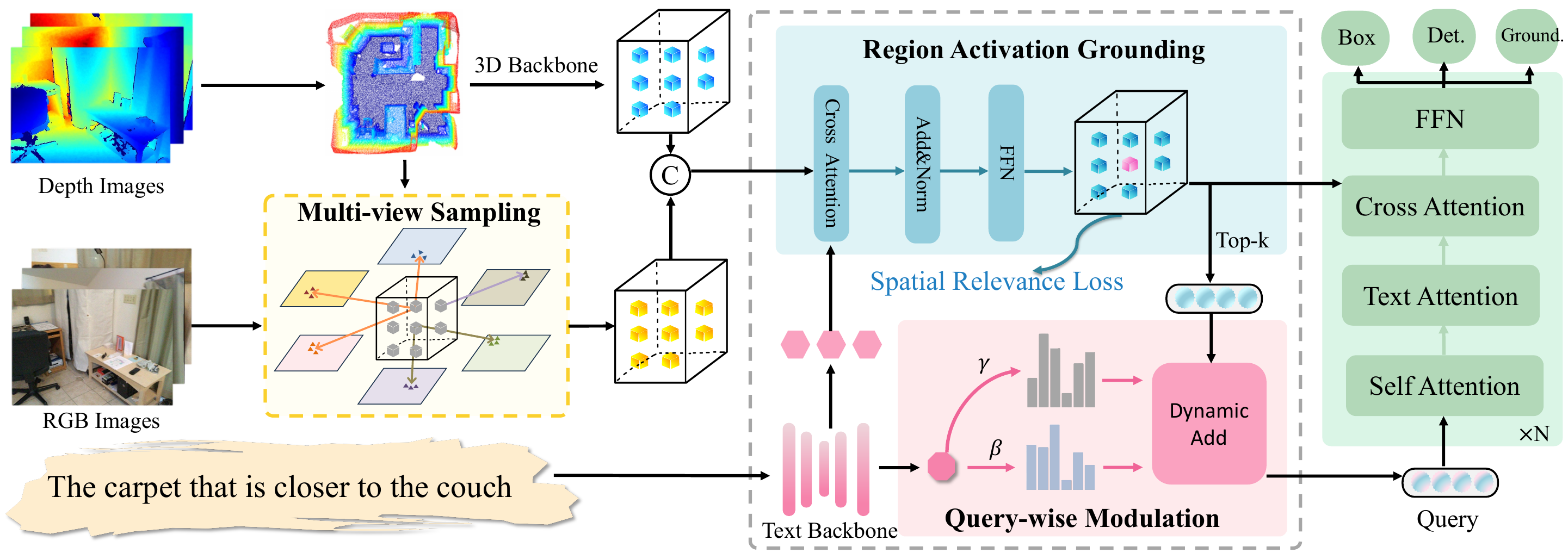}
	\vspace{-14pt}
	\caption{\textbf{The overall framework of our DEGround}.
    It shares DETR queries as the object representation for both detection and grounding.
    Besides, it includes a Region Activation Grounding (RAG) module for highlighting instruction-relevant regions and a Query-wise Modulation (QIM) module that dynamically modulates decoder queries with global linguistic context.}
	\label{fig:framework}
	\vspace{-8pt}
\end{figure*} 
In this work, we introduce DEGround, designed to maximize the utilization of pre-trained detection capabilities. By sharing query across both tasks, DEGround avoids disjoint representational spaces and redundant learning
(\S\ref{sec:framework}).
Additionally, for grounding we introduce two modules. RAG (\S\ref{sec:fusion}) highlights regions relevant to textual cues, and QIM (\S\ref{sec:adapter}) injects sentence level semantics into the queries.
The overall framework is shown in Fig.~\ref{fig:framework}.
In the following sections, we will delve into more model details.

\subsection{Overall Framework}
\label{sec:framework}
Both detection and grounding take multiple RGB-D image sequences as input and output 3D bounding boxes. The two tasks use the same feature pipeline to extract geometric and semantic cues, which are fused into a unified 3D feature representation. With shared DETR-style object queries, a single decoder operates on this representation to predict 3D boxes for both tasks. For grounding, the queries are additionally conditioned on the referring expression.

\noindent\textbf{Feature Extraction.}
Given $V$ views of RGB-D inputs $\{(\bm{I}_v,\bm{D}_v)\}^V_{v=1}$, where each $\bm{I}_v$ represents an RGB image, and $ \bm{D}_v$ corresponds to its depth map.
We first reconstruct a sparse 3D scene representation by projecting the multi-view depth maps $\{\bm{D}_v\}^V_{v=1}$ into world coordinates using their corresponding depth sensor intrinsic and extrinsic parameters. This process generates a pseudo-point cloud that preserves geometric fidelity to the physical environment. The pseudo-point cloud is then voxelized and encoded through a voxel encoder, producing multi-level 3D features $\bm{F}_{3d}\! \in\! \mathbb{R}^{N\times C} $ and their 3D coordinates $\bm{F}_{coord}\!\in\! \mathbb{R}^{N\times 3}$. 
Here, $N$ is the voxel number, $C$ represents the feature dimension, and we omit the level notion for simplicity.
Concurrently, an image encoder is used to extract multi-scale 2D features from each RGB image, obtaining corresponding feature maps $\bm{F}_{2d}\! \in\! \mathbb{R}^{H\times W \times C'}$ per view.
Let $\bm{L}$ denote a language description with $T$ words. 
We employ a pre-trained text backbone to extract text embeddings $\bm{F}_{text}\! \in\! \mathbb{R}^{T \times C''}$.
These three types of features are projected into the same feature dimension $C$ via independent linear layers.

\noindent\textbf{3D Feature Representation.}
Following the projection of all visual features into a unified embedding space, we construct a comprehensive 3D feature representation by hierarchically integrating two complementary feature streams: (i) 3D features $\bm{F}_{3d}$, which encode the geometric structure of the scene, and (ii) 2D features $\bm{F}_{2d}$, which provide rich semantic context from the RGB modalities. Here, each $\bm{F}_{2d}$ incorporates multi-view features, \ie $\bm{F}_{2d} = \{\bm{F}_{v,2d} \}_{v=1}^V$. 

To align 2D and 3D features, we first convert voxel indices to 3D point coordinates $\bm{P}\!\in\! \mathbb{R}^{N\times 3}$ using the voxel size. These points are then projected onto multi-view image planes using camera parameters, and their corresponding 2D features are sampled via bilinear interpolation. Finally, the interpolated 2D features are concatenated with the 3D geometric features along the channel dimension, resulting in a fused visual representation $\bm{F}_{visual}$ that integrates both semantic and spatial information:
\begin{equation}
    \bm{F}_{visual} = \text{Concat}\left(\bm{F}_{3d} , Sample(\bm{P},\bm{F}_{2d})\right)\! \in\! \mathbb{R}^{N\times C}.
\end{equation}

\noindent\textbf{Sharing Query.}
Different from previous works~\cite{wang2024embodiedscan,zhengdensegrounding,peng2025proxytransformation} that use two distinct decoder architectures, with detection relying on a convolution-based approach and grounding employing a transformer-based method, we use a set of queries as the common object representation for both tasks. 
Concretely, a lightweight scoring head predicts task-specific confidence scores for each voxel center in $\bm{F}_{coord}$, based on its corresponding visual feature in $\bm{F}_{visual}$.
The predicted scores, representing detection classification logits or grounding confidence values, are ranked, and the top-K highest scoring candidates are selected as the initial query positions.
The corresponding feature vectors are then extracted and used to form the query embeddings $\bm{Q}$.
The shared decoder $\mathcal{F}^{Dec}$, built upon the standard Transformer architecture~\cite{carion2020end}, processes $\bm{F}_{visual}$, $\bm{F}_{text}$, and $\bm{Q}$ to predict 3D bounding boxes and corresponding scores:
\begin{equation}
    \bm{B}, \bm{C}^{det}, \bm{C}^{grd} = \mathcal{F}^{Dec}(\bm{F}_{visual}, \bm{F}_{text}, \bm{Q}).
\end{equation}

Each $\bm{B}$ is a set of 9-degree-of-freedom (9DoF) bounding boxes $(x, y, z, l, w, h, \alpha, \beta, \gamma)$. Here, $(x,y,z)$ are the 3D coordinates of an object, $(l,w,h)$ are its dimensions, and $(\alpha,\beta,\gamma)$ represents its orientation angles.
$\bm{C}^{det}\in\mathbb{R}^{K\times C^{det}}$ and $\bm{C}^{grd}\in\mathbb{R}^{K\times C^{grd}}$ denote the final detection and grounding classification scores, respectively.
Both branches share the same decoder but use independent MLP heads.
This sharing query mechanism enables the more challenging grounding task to benefit from the robust localization and categorization priors learned by detection.

\subsection{Regional Activation Grounding}
\label{sec:fusion}
A significant challenge in ego-centric 3d visual grounding is same-class distractors, which share the target’s category but are inconsistent with the language description. To overcome this, we propose the Regional Activation Grounding (RAG) module, which explicitly highlights text-consistent regions.
RAG consists of two key components: Text-driven Attention for cross-modal feature interaction and Spatial Relevance Loss to supervise region-level grounding. Details are provided in the following sections.

\noindent\textbf{Text-driven Attention.}
Formally, we first employ a text-driven attention module $\mathcal{F}^{att}$ to fuse the 3D visual feature $\bm{F}_{visual}$ with the linguistic representation $\bm{F}_{text}$. This module computes attention weights based on the semantic relevance between the two modalities, allowing the model to highlight regions in the visual space that are most relevant to the given instruction, denoted as $\bm{F}_{region}$:
\begin{equation}
    \bm{F}_{region}=\mathcal{F}^{att}(Q\!=\!\bm{F}_{visual},K\!=\!\bm{F}_{text},V\!=\!\bm{F}_{text}).
\end{equation}
To preserve the original spatial information while enriching it with linguistic context, we adopt a residual connection by adding $\bm{F}_{region}$ back to the original 3D visual features:
\begin{equation}
    \bm{F}_{region}' = \bm{F}_{visual} + \bm{F}_{region} \! \in\! \mathbb{R}^{N\times C},
\end{equation}
where the final text-enhanced visual features $\bm{F}_{region}'$ can be a seamless insertion into the above overall framework.
Finally, we will introduce an auxiliary spatial relevance loss to explicitly activate these related regions, further improving the grounding accuracy.

\noindent\textbf{Spatial Relevance Loss.} 
Specifically, a multi-layer perceptron (MLP) is employed to predict a spatial relevance score $\bm{S} \in \mathbb{R}^{N \times 1}$ for each 3D point based on the fused feature representation $\bm{F}_{\text{region}}'$.
\begin{equation}
    \bm{S} = \mathcal{F}^{MLP}(\bm{F}_{region}')  \! \in\! \mathbb{R}^{N\times 1}.
\end{equation}
The spatial relevance score represents the likelihood of a point belonging to a visual region associated with the given text.
To supervise this prediction, we derive ground-truth labels $\hat{\bm{S}} \in \mathbb{R}^{N \times 1}$ based on the annotated 3D bounding box: points located within the ground-truth region are assigned a label of 1, while those outside are assigned 0.
The spatial relevance loss $\mathcal{L}_{spatial}$ is then computed using cross-entropy loss, comparing the predicted regional similarity scores $\bm{S}$ with these binary labels $\hat{\bm{S}}$. 
By performing regional activation grounding, the text-aware regions are effectively activated.
In the following section, we extend this idea to query-level activation with context-aware modulation.

\subsection{Query-wise Modulation}
\label{sec:adapter}
While RAG facilitates global alignment between linguistic and visual modalities, it exhibits limited adaptability in modulating decoder queries according to task-specific semantic requirements during grounding. To address this limitation, we propose a Query-wise Modulation module that infuses linguistic cues into visual decoder queries via a lightweight yet expressive modulation mechanism.

As shown in ~\cref{fig:framework}, the core idea is to dynamically reshape the representation space of each decoder query according to the input instruction, thereby enhancing its semantic specificity and grounding capacity. 
Concretely, given a sentence embedding $\bm{S} \in \mathbb{R}^{C}$, we employ two independent multi-layer perceptrons (MLPs), denoted as $\xi_1$ and $\xi_2$, to generate modulation vectors $\bm{\beta}, \bm{\gamma} \in \mathbb{R}^{C}$:
\begin{equation}
\begin{split}
    \bm{\beta} = \xi_1(\bm{S}), \\
\bm{\gamma} = \xi_2(\bm{S}).
\end{split}
\end{equation}
These vectors encode the attention-to-vision and attention-to-language balance, respectively. Given an init set of $K$ visual decoder queries $\bm{Q}_{vis} \in \mathbb{R}^{K \times C}$, the final language-aware queries $\bm{Q}_{mod} \in \mathbb{R}^{K \times C}$ are obtained via a feature-wise affine transformation:
\begin{equation}
\bm{Q}_{mod} = \bm{\beta} \odot \bm{Q}_{vis} + \bm{\gamma} \odot \bm{1}_M \bm{S},
\end{equation}

where $\odot$ denotes element-wise multiplication and $\bm{1}_M \in \mathbb{R}^{M \times 1}$ is a broadcast vector that replicates $\bm{S}$ across all queries. This formulation enables each decoder query to be adaptively modulated by both visual and linguistic signals, thereby promoting more effective context-aware grounding.

\subsection{Loss Function}
We now introduce the loss function used to train our proposed DEGround in this part.
To establish correspondences between predictions and ground-truth objects, we first find a bipartite graph matching which of the predicted objects fits the ground-truth.
We search for a permutation of predictions by minimizing matching cost. 
The optimal assignment is then used to compute the detection loss:
\begin{equation}
    \mathcal{L}_{det} = \lambda_{cls}\mathcal{L}_{cls} + \lambda_{box}\mathcal{L}_{box},
\end{equation}
where $\mathcal{L}_{cls}$ is the class-related loss implemented via focal loss~\cite{lin2017focal}, and $\mathcal{L}_{box}$ is the box regression loss. $\lambda_{cls}$ and $\lambda_{box}$ are the corresponding hyper-parameters.

The same bipartite matching process is applied to align the predicted grounding boxes with the ground-truth annotations. 
The grounding loss includes three components: a focal classification loss $\mathcal{L}_{ground}$, a box regression loss $\mathcal{L}_{box}$, and a spatial relevance loss $\mathcal{L}_{spatial}$ introduced in \S\ref{sec:fusion}. The full grounding objective is defined as:
\begin{equation}
    \mathcal{L}_{ground}= \lambda_{ground}\mathcal{L}_{ground} + \lambda_{box}\mathcal{L}_{box} + \lambda_{spatial}\mathcal{L}_{spatial}.
\end{equation}

\section{Experiments}
\label{sec:experiments}
\begin{table*}[t]
\caption{\textbf{Comparison with state-of-the-art methods on EmbodiedScan multi-view 3D detection benchmark.}
The overall precision score is reported on three sub-datasets: ScanNet, 3RScan, and MP3D.
`MP3D' is the Matterport3D sub-dataset in EmbodiedScan.
}
\vspace{-3mm}
\centering
\setlength{\tabcolsep}{9pt}
\resizebox{\linewidth}{!}{
\begin{tabular}{l|c|ccc|ccc|ccc}
\toprule
\rowcolor[gray]{.9}

Method  &Overall  &Head &Common &Tail &Small  &Medium &Large &ScanNet &3RScan &MP3D \\
\midrule
VoteNet~\cite{ding2019votenet} &5.18 &10.87 &2.41 &2.07 &0.16 &5.30 &5.99 &9.90 &7.69 &3.82 \\
\rowcolor{StripeGray}ImVoxelNet~\cite{rukhovich2022imvoxelnet} &8.08 &3.11 &7.05 &3.73 &0.06 &7.95 &9.02 &11.91 &2.17 &5.24 \\
FCAF3D~\cite{rukhovich2022fcaf3d} &13.86 &22.89 &9.61 &8.75 &2.90 &13.90 &10.91 &21.35 &17.02 &9.78 \\
\rowcolor{StripeGray}EmbodiedScan~\cite{wang2024embodiedscan}  &15.22 &24.95 &10.81 &9.48 &3.28 &15.24 &10.95 &22.66 &18.25 &10.91 \\
BIP3D~\cite{lin2024bip3d} &20.91 &27.57 &18.77 &16.03 &5.72 &21.48 &15.20 &23.47 &32.48 &10.09 \\
\rowcolor{cyan!10}DEGround (Ours)  &\textbf{24.68} &\textbf{34.45} &\textbf{19.71} &\textbf{19.60} &\textbf{9.23} &\textbf{24.94} &\textbf{15.61} &\textbf{23.62} &\textbf{39.53} &\textbf{10.78} \\

\bottomrule
\end{tabular} 
}
\label{tab:detection}
\end{table*}

\begin{table*}[t]
\caption{\textbf{Comparison with state-of-the-art methods on EmbodiedScan visual grounding benchmark.} $^\dagger$ denotes the result reproduced in our experiments. 
`-' means unavailable.
The overall precision score is also reported on three sub-datasets: ScanNet, 3RScan, and MP3D.
`MP3D' refers to the Matterport3D sub-dataset in EmbodiedScan.
The best results are in bold.}
\vspace{-3mm}
\centering
\resizebox{\linewidth}{!}{
\begin{tabular}{l|l|c|c|cc|cc|ccc}
\toprule
\rowcolor[gray]{.9}
\cellcolor[gray]{.9}Set & Method & Backbone & Overall  &Easy &Hard &View-Dep &View-Indep  & ScanNet & 3RScan & MP3D\\

\midrule
\multirow{4}{*}{\makecell{Mini}}
&EmbodiedScan$^\dagger$~\cite{wang2024embodiedscan}& ResNet-50 &35.84 &36.28 &30.81 &36.56 &35.46 &40.00 &35.26 &29.34 \\
&\cellcolor{StripeGray}DenseGrounding~\cite{zhengdensegrounding}&\cellcolor{StripeGray}ResNet-50&\cellcolor{StripeGray}41.34 &\cellcolor{StripeGray}41.95&\cellcolor{StripeGray}34.38&\cellcolor{StripeGray}40.89&\cellcolor{StripeGray}42.19&\cellcolor{StripeGray}44.45&\cellcolor{StripeGray}41.08&\cellcolor{StripeGray}33.72 \\
&BIP3D~\cite{lin2024bip3d}&Swin-T &45.79 &46.22 &40.91 &45.93 &45.71 &48.94 &46.61 &37.36 \\
&\cellcolor{cyan!10}DEGround (Ours)& \cellcolor{cyan!10}ResNet-50& \cellcolor{cyan!10}\textbf{61.28} &\cellcolor{cyan!10}\textbf{61.76} &\cellcolor{cyan!10}\textbf{55.84} &\cellcolor{cyan!10}\textbf{62.95} &\cellcolor{cyan!10}\textbf{60.39} &\cellcolor{cyan!10}\textbf{62.71} &\cellcolor{cyan!10}\textbf{65.03} &\cellcolor{cyan!10}\textbf{51.65} \\

\midrule
\multirow{8}{*}{\makecell{Full}}
&ScanRefer~\cite{chen2020scanrefer} &- &12.85 &13.78 &9.12 &13.44 &10.77 &- &- &- \\
&\cellcolor{StripeGray}BUTD-DETR~\cite{jain2022bottom} &\cellcolor{StripeGray}- &\cellcolor{StripeGray}22.14 &\cellcolor{StripeGray}23.12 &\cellcolor{StripeGray}18.23 &\cellcolor{StripeGray}22.47 &\cellcolor{StripeGray}20.98 &\cellcolor{StripeGray}- &\cellcolor{StripeGray}- &\cellcolor{StripeGray}- \\
&L3Det~\cite{zhu2023object2scene} &- &23.07 &24.01 &18.34 &23.59 &21.22 &- &- &- \\
&\cellcolor{StripeGray}EmbodiedScan~\cite{wang2024embodiedscan} &\cellcolor{StripeGray}ResNet-50 &\cellcolor{StripeGray}39.41 &\cellcolor{StripeGray}40.12 &\cellcolor{StripeGray}31.45 &\cellcolor{StripeGray}40.21 &\cellcolor{StripeGray}38.96 &\cellcolor{StripeGray}41.99 &\cellcolor{StripeGray}41.53 &\cellcolor{StripeGray}30.29 \\
&ProxyTransformation~\cite{peng2025proxytransformation}&ResNet-50 &41.08 &41.66 &34.38 &41.57 &40.81 &- &- &- \\
&\cellcolor{StripeGray}DenseGrounding~\cite{zhengdensegrounding} &\cellcolor{StripeGray}ResNet-50 &\cellcolor{StripeGray}44.41 &\cellcolor{StripeGray}45.31 &\cellcolor{StripeGray}34.23 &\cellcolor{StripeGray}44.42 &\cellcolor{StripeGray}44.40 &\cellcolor{StripeGray}- &\cellcolor{StripeGray}- &\cellcolor{StripeGray}- \\
&BIP3D~\cite{lin2024bip3d} &Swin-T &54.66 &55.07 &50.12 &55.78 &54.03 &61.23 &55.41 &39.36 \\
&\cellcolor{cyan!10}DEGround (Ours)& \cellcolor{cyan!10}ResNet-50&\cellcolor{cyan!10}\textbf{62.18} &\cellcolor{cyan!10}\textbf{62.76} &\cellcolor{cyan!10}\textbf{55.70} &\cellcolor{cyan!10}\textbf{63.56} &\cellcolor{cyan!10}\textbf{61.40} &\cellcolor{cyan!10}\textbf{63.02} &\cellcolor{cyan!10}\textbf{65.98} &\cellcolor{cyan!10}\textbf{52.95} \\

\bottomrule
\end{tabular} 
}
\label{tab:grounding}
\end{table*}

\begin{table}[t]
\caption{\textbf{Comparison with state-of-the-art methods} on EmbodiedScan grounding \textbf{test} set.}
\vspace{-3mm}
\centering
\resizebox{\linewidth}{!}{
\begin{tabular}{l|c|c|cc}
\toprule
\rowcolor[gray]{.9}
Method &Backbone  & \makecell{Overall\\ $\mathrm{AP}_{50}$}&\makecell{Easy\\ $\mathrm{AP}_{50}$} &\makecell{Hard\\ $\mathrm{AP}_{50}$}\\

\midrule
EmbodiedScan~\cite{wang2024embodiedscan}&ResNet-50 &16.35&16.71&12.37 \\
\rowcolor{StripeGray}SAG3d~\cite{liang2024spatioawaregrouding3d} &ResNet-50 &20.38&20.91&14.49 \\
DenseGrounding~\cite{zhengdensegrounding}&ResNet-50 &34.72 &35.46&26.56 \\
\rowcolor{StripeGray}BIP3D~\cite{lin2024bip3d}&Swin-T &39.69 &40.40&31.77 \\
\cellcolor{cyan!10}DEGround (Ours)&\cellcolor{cyan!10}ResNet-50 &\cellcolor{cyan!10}\textbf{42.04} &\cellcolor{cyan!10}\textbf{42.65} &\cellcolor{cyan!10}\textbf{35.18} \\

\bottomrule
\end{tabular} 
\vspace{-8pt}
}
\label{tab:supply_grounding}
\vspace{-2mm}
\end{table}

\begin{table*}[t!]
\caption{\small \textbf{Ablation studies of different architecture designs.} We evaluate how hyperparameter settings influence model performance.}
\centering
\vspace{-3mm}
\footnotesize
\setlength{\tabcolsep}{8.5pt} 
\begin{tabular}{@{}c@{\hspace{0.02\linewidth}}c@{\hspace{0.02\linewidth}}c@{}}
  \begin{subtable}[t]{0.32\linewidth}
    \caption{\small Different query numbers for grounding.}
    \label{table:grounding_query_number}
    \centering
    \begin{tabular}{c|ccc}
      \rowcolor[gray]{.9}
      \toprule
      Number & Overall & Easy & Hard \\
      \midrule
      100  & 59.52 & 59.97 & 54.36 \\
      256  & 60.53 & 60.91 & \textbf{56.15} \\
      \rowcolor{cyan!10}512 & \textbf{61.28} & \textbf{61.76} & 55.84 \\
      \bottomrule
    \end{tabular}
  \end{subtable}
  &
  \begin{subtable}[t]{0.32\linewidth}
    \caption{\small Different spatial relevance loss weights.}
    \label{table:loss_weight}
    \centering
    \begin{tabular}{c|ccc}
      \rowcolor[gray]{.9}
      \toprule
      Weight & Overall & Easy & Hard \\
      \midrule
      0.005 & 60.80 & 61.18 & 56.47 \\
      \rowcolor{cyan!10}0.01 & \textbf{61.28} & \textbf{61.76} & 55.84 \\
      0.05  & 60.70 & 60.96 & \textbf{57.73} \\
      \bottomrule
    \end{tabular}
  \end{subtable}
  &
  \begin{subtable}[t]{0.32\linewidth}
    \caption{\small Different query numbers for detection.}
    \label{table:detection_query_number}
    \centering
    \begin{tabular}{c|ccc}
      \rowcolor[gray]{.9}
      \toprule
      Number & Overall & Easy & Hard \\
      \midrule
      256   & 22.48 & 32.11 & 19.19 \\
      512   & 23.20 & 33.13 & 19.30 \\
      \rowcolor{cyan!10}1024 & \textbf{24.68} & \textbf{34.45} & \textbf{19.71} \\
      \bottomrule
    \end{tabular}
  \end{subtable}
\end{tabular}
\end{table*}

\subsection{Dataset and Evaluation Metric}
\label{sec:dataset_evaluation}

\noindent\textbf{Dataset.}
Following~\cite{zhengdensegrounding}, we evaluate our method on EmbodiedScan~\cite{wang2024embodiedscan}, a large-scale, multi-modal, ego-centric 3D perception benchmark. 
It is composed of 4,633 scans from three well-known datasets: ScanNet ~\cite{scannet}, 3RScan~\cite{wald2019rio}, and Matterport3D~\cite{chang2017matterport3d}.
Instead of relying on reconstructed scene-level points or meshes, EmbodiedScan uses first-person RGB-D streams that capture realistic embodied trajectories and viewpoint changes.
It provides 3D bounding boxes for 284 object categories and approximately 970k language descriptions.
This extensive dataset provides a diverse and rich foundation for 3D visual grounding tasks, offering broader scene coverage compared to prior datasets. This makes the dataset significantly larger and more challenging than previous ones, providing a more rigorous benchmark for 3D visual grounding tasks.

\noindent\textbf{Evaluation Metric.}
For object detection, we follow the standard EmbodiedScan protocol~\cite{wang2024embodiedscan} and report 3D IoU-based Average Precision at a 0.25 threshold (AP@25).
We further report performance across three dimensions: category generalization (head, common, and tail categories), object scale (small, medium, and large sizes), and scene-level generalization (across different scene types).
For the 3D grounding task, AP@25 also serves as the main evaluation metric. We additionally analyze performance with respect to instruction difficulty (based on the number of distractor objects), view dependency (whether the instruction contains directional cues), and sub-scene variations. 
For submissions to the official test server, we follow the leaderboard protocol and report AP@50, a more challenging metric requiring higher localization precision.

\subsection{Implementation Details}
\label{sec:implementation}
\noindent\textbf{Model.} For feature extraction, we adopt ResNet50~\cite{he2016deep} for 2D semantic features,  Minkowski ResNet34~\cite{choy20194d} for 3D geometric features, and RoBERTa~\cite{liu2019roberta} for textual features as the respective backbones.
The decoder is composed of 6 Transformer layers, while the box, classification, and grounding heads, built on top of the decoder.

\noindent\textbf{Training.}
We adopt the AdamW optimizer~\cite{loshchilov2017decoupled} for network training.
Data augmentation includes random flipping and random rotation in 3D space. 
For 3D detection, both loss weights $\lambda_{cls}$ and $\lambda_{box}$ are set to 1.
For 3D grounding, the loss weights are set as: $\lambda_{ground} = 1$, $\lambda_{box} = 1$, and $\lambda_{spatial} = 0.01$.
More implementation details can be found in the supplementary material.

\subsection{Comparison with State-of-The-Art}
\label{sec:sota}
\noindent\textbf{3D Detection.} In Table \ref{tab:detection}, we report results on EmbodiedScan visual detection benchmark. Previous methods usually rely on convolutional architectures due to the concerns about convergence challenges of transformer-based methods on large-scale datasets with over 200 object categories. Contrary to this assumption, our DETR-based framework achieves 24.68\% AP@25 on the overall dataset, which is 3.77\% higher than the previous state-of-the-art BIP3d~\cite{lin2024bip3d}. The improvements range from 20.91\% to 24.68\%, demonstrating the effectiveness of our approach.

\noindent\textbf{3D Grounding.} We compare our model DEGround with existing methods~\cite{chen2020scanrefer, jain2022bottom, zhu2023object2scene, wang2024embodiedscan, peng2025proxytransformation,zhengdensegrounding,lin2024bip3d} on the EmbodiedScan visual grounding benchmark. The results are shown in Table~\ref{tab:grounding}.
Remarkably, DEGround outperforms all methods on both the full and mini validation datasets under all metrics, which demonstrates the superior effectiveness and robustness of our method. 
In specific, 
on the full validation dataset, DEGround with a ResNet-50 backbone achieves an AP of \textbf{62.18}\%. Compared to other models that employ ResNet-50 as the backbone, DEGround surpasses the previous state-of-the-art DenseGrounding~\cite{zhengdensegrounding} by a large margin of \textbf{17.77}\%. Furthermore, even compared with BIP3D~\cite{lin2024bip3d}, which adopts a stronger Swin-T backbone, DEGround still achieves a \textbf{7.52\%} improvement (62.18 vs. 54.66). 
On the mini validation dataset, DEGround maintains a consistent advantage with an AP of 61.28\%, which is 15.49\% higher than the previous state-of-the-art BIP3D~\cite{lin2024bip3d}. It also converges faster, reaching 61.28\% in 3 epochs, while our baseline reaches 35.84\% after 12 epochs. These results show that the sharing query design unifies the object representation across detection and grounding, enabling effective object-level knowledge transfer and improving both accuracy and data efficiency.

We further compare our DEGround with existing methods on the EmbodiedScan visual grounding test set, which uses a non-public test set maintained by the official benchmark. 
As shown in Tab.~\ref{tab:supply_grounding}, DEGround achieves state-of-the-art performance, outperforms the nearest
competitor BIP3d~\cite{lin2024bip3d} by 2.35\% AP@50 on overall precision.
\subsection{Ablation Studies}
\label{ablation}
To offer a deep insight into our DEGround, we conduct ablation studies to analyze the effectiveness of each component. If not specialized, we report our performance on the EmbodiedScan mini validation dataset, with 1024 and 512 queries for detection and grounding, respectively.

\noindent\textbf{Module Effectiveness.} We conduct ablation experiments to evaluate the effectiveness of different components. As shown in Tab.~\ref{tab:comp_abl}, introducing the sharing query leads to a remarkable improvement of 24.06\% AP@25 over the vanilla baseline, which is adopted from EmbodiedScan~\cite{wang2024embodiedscan}. These notable improvement demonstrates the effectiveness of our sharing-query mechanism. Even on this strong baseline, adding RAG and QIM further enhances performance, particularly on the hard subset, with additional gains of +1.79\% and +1.16\%, respectively. With all components integrated, the model attains the best result of 61.28\%.
\begin{figure*}[t]
\centering
\includegraphics[width=\linewidth]{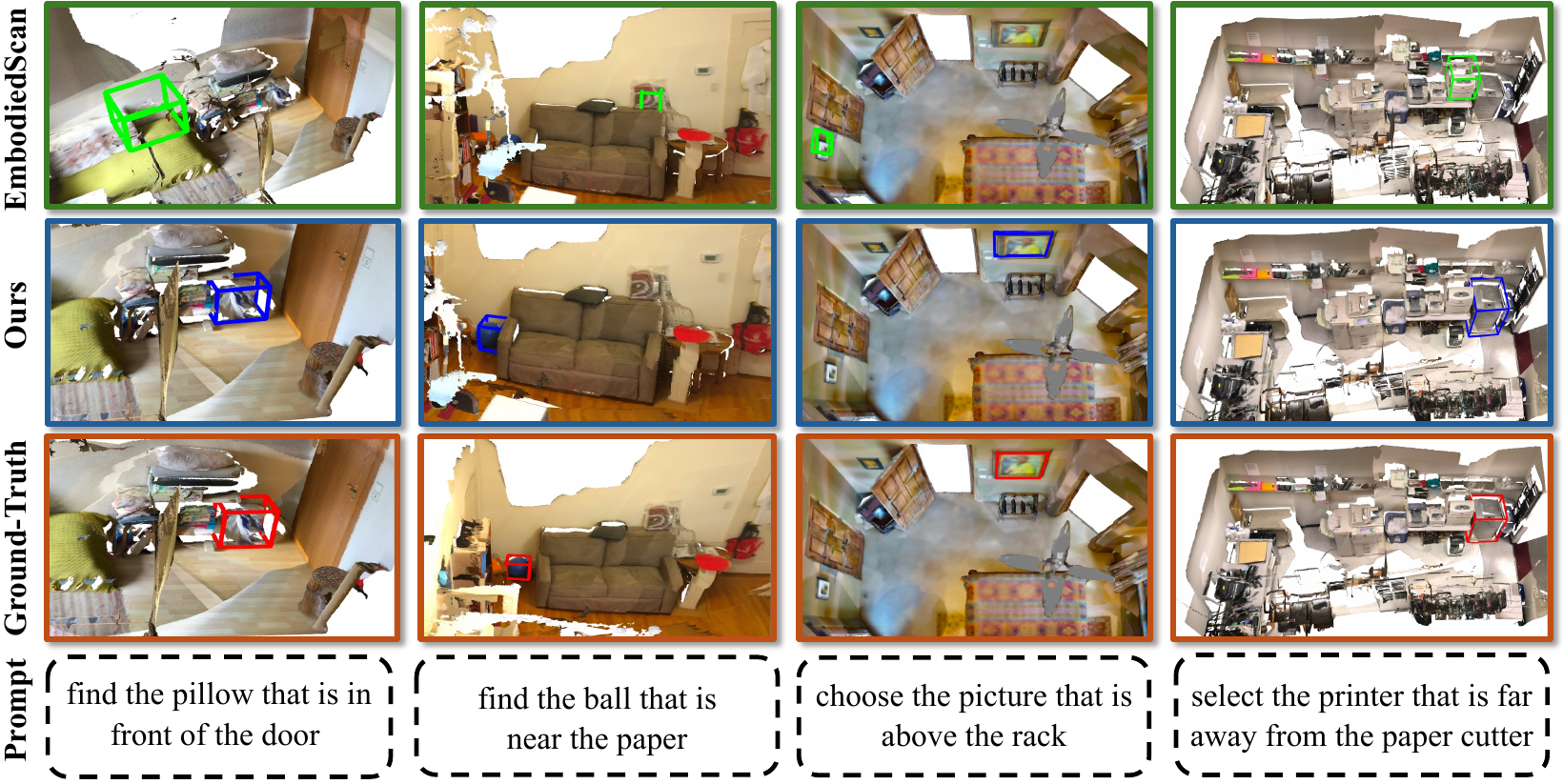}
\vspace{-16pt}
\caption{\textbf{Qualitative comparisons of our method and EmbodiedScan.} Our method shows stronger context-aware grounding and is notably more robust to same-category distractors.
} 
\label{fig:visible}
\vspace{-10pt}
\end{figure*} 

\begin{table}[t]
\caption{\textbf{Ablation study of our method.} Shar Q. uses sharing query to unify object representations for detection and grounding.
}
\centering
\setlength{\tabcolsep}{6pt}
\resizebox{0.8\linewidth}{!}{
\begin{tabular}{ccc|ccc}
\toprule
\rowcolor[gray]{.9} Shar Q. & RAG &QIM &Overall &Easy &Hard \\
\midrule
-&-&-&35.84&36.28&30.81 \\
\ding{52}&-&-&59.90&60.37&54.57 \\
\ding{52}&\ding{52}&-&61.10&61.52&\textbf{56.36} \\
\ding{52}&-&\ding{52}&60.63&61.06&55.73 \\
\rowcolor{cyan!10}\ding{52}&\ding{52}&\ding{52}&\textbf{61.28} &\textbf{61.76} &55.84 \\
\bottomrule
\end{tabular} 
}
\vspace{-8pt}
\label{tab:comp_abl}
\end{table}

\noindent\textbf{Grounding query numbers.} It is also of interest to explore the impact of different query numbers. As shown in Tab.~\ref{table:grounding_query_number}, for the grounding task, increasing the number of queries leads to consistent performance improvements, with AP@25 gains of 1.01\% and 0.75\% observed when increasing the number of queries from 100 to 256 and 512, respectively. 
Considering that increasing the number of queries also increases computational resource consumption, we adopt 512 queries during training to achieve a favorable trade-off between accuracy and computational efficiency.

\noindent\textbf{About spatial relevance loss.} The spatial relevance loss is designed to refine the model’s ability to localize target regions. We explore different weights for this loss component to determine its impact on our model. As shown in Tab.~\ref{table:loss_weight}, a weighting of 0.01 achieves the optimal results of 61.28\%, outperforming both lower (0.005) and higher (0.05) weighting configurations. Based on these results, we set the region loss weight as $\lambda_{spatial} = 0.01$ throughout our experiments.

\noindent\textbf{Detection query numbers.}
We further examine the effect of varying the number of detection queries on detection performance in Tab.~\ref{table:detection_query_number}. Specifically, we evaluate our model with 256, 512, and 1024 queries. The detection accuracy improves progressively from 22.48\% with 256 queries to 23.20\% with 512 queries, and reaches 24.68\% with 1024 queries. 
This trend indicates that increasing the number of detection queries enhances the model's capacity to capture diverse object instances and improves its ability to localize targets more precisely. 
However, this improvement also increases computational resource consumption.
Finally, we adopt 1024 queries for all subsequent experiments to balance performance and efficiency.

\subsection{Qualitative Results}
We provide qualitative comparisons between our method and EmbodiedScan~\cite{wang2024embodiedscan}, as shown in Fig.~\ref{fig:visible}. Across various scenarios, our approach demonstrates superior grounding capability. For example, given the prompt ``choose the picture that is
above the rack'', EmbodiedScan is confused by same-class distractors, whereas our method accurately identifies the referred cabinet, demonstrating stronger discriminative ability.
Moreover, in the last column with densely cluttered scenes, our method still precisely locates the target object, highlighting its robustness and reliability. Additional results are provided in the supplementary material.

\begin{figure}[t]
        \centering
        \includegraphics[width=\linewidth]{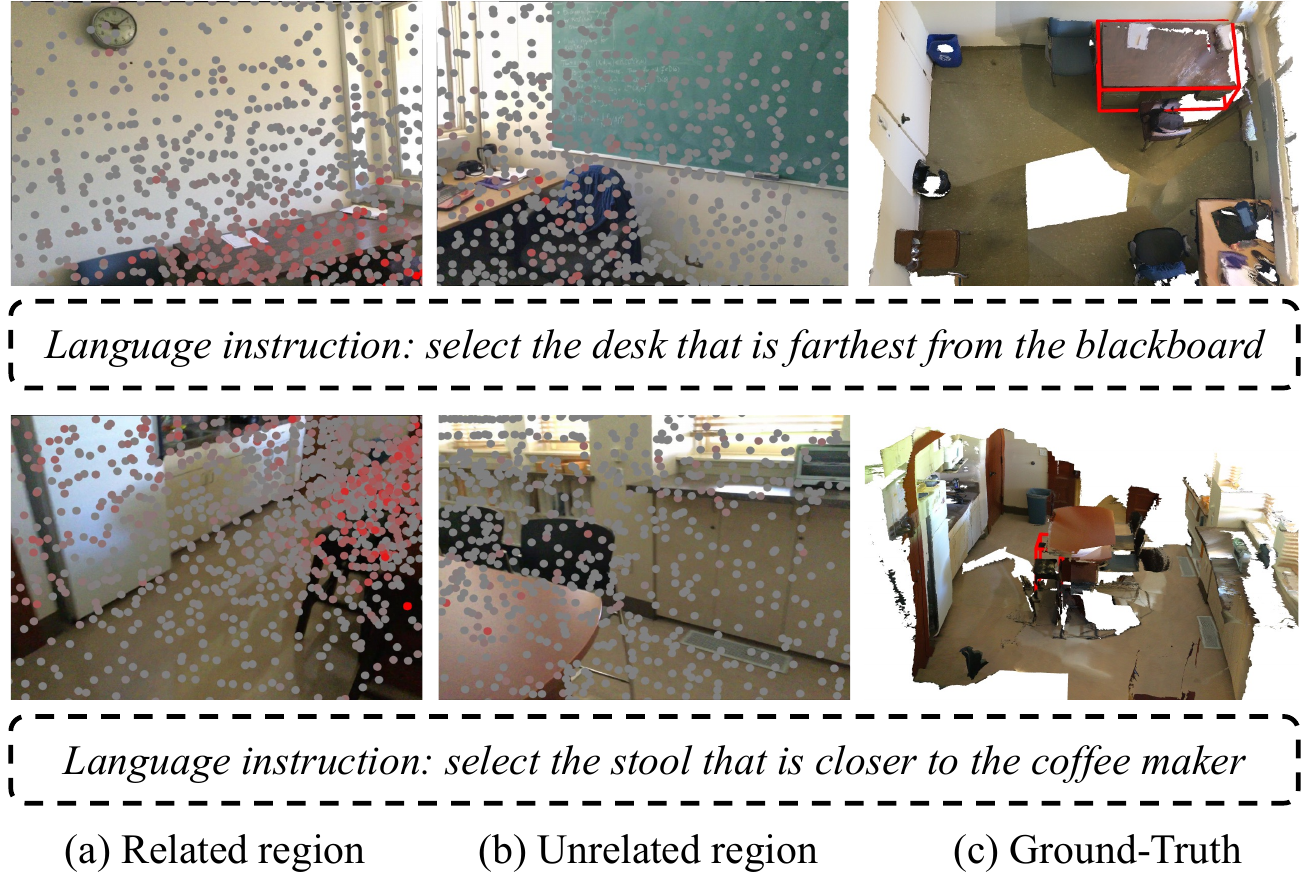}
    \vspace{-15pt}
    \caption{\textbf{Viusalization of regional similarity scores in RAG.} 
    RAG highlights higher relevant regions.
    } 
    \label{fig:variance}
    \vspace{-10pt}
\end{figure} 

In Fig.~\ref{fig:variance}, we also visualize the regional similarity score generated by RAG. We first project these scores onto RGB images from different perspectives. 
Then render color-coded heatmaps according to the score value, where gray represents low similarity and red represents high similarity. 
The visualization clearly demonstrates that the RAG module effectively highlights regions with higher text similarity.

\section{Conclusion}
In this paper, we presented DEGround, a homogeneous framework that unifies 3D detection and grounding object representation through sharing object queries. By aligning both tasks within a common query space, DEGround enables efficient object-level knowledge transfer and eliminates redundant optimization between detection and grounding. 
To further enhance grounding performance, we introduce two task-specific modules that rewrite the visual feature map and query space, respectively. 
RAG refines the visual feature map, mitigating same-class distractor confusion by strengthening spatial–textual correspondence, while QIM applies sentence-conditioned affine modulation to initialize instruction-aware query embeddings.
Extensive experiments demonstrate that DEGround consistently outperforms existing methods, achieving state-of-the-art results across both validation and test splits.

{
    \small
    \bibliographystyle{ieeenat_fullname}
    \bibliography{main}
}


\end{document}